# Analysis in HUGIN of Data Conflict

Finn Verner Jensen, Bo Chamberlain, Torsten Nordahl, Frank Jensen,

Aalborg University, Department of Mathematics and Computer Science
Fredrik Bajers Vej 7, DK-9220 Aalborg, Denmark

June 8, 1990

## Abstract

After a brief introduction to causal probabilistic networks and the HUGIN approach, the problem of conflicting data is discussed. A measure of conflict is defined, and it is used in the medical diagnostic system MUNIN. Finally it is discussed how to distinguish between conflicting data and a rare case.

## 1   Introduction

It has for many years been widely recognized that causal probabilistic networks (CPN's), have many virtues with respect to expert systems mainly due to the transparency of the knowledge embedded and their ability to unify almost all domain knowledge relevant for an expert system (Pearl 1988). However, the calculation of revised probability distributions after the arrival of new evidence was for a long period intractable and therefore an obstacle for pursuing these virtues. Theoretical developments in the 80ies have overcome this difficulty (Kim and Pearl 1983, Lauritzen and Spiegelhalter 1988, Schachter 1988, Cooper 1984, Shafer and Shenoy 1989). The Lauritzen and Spiegelhalter method has been further developed to the HUGIN approach (Andersen et al. 1987, Jensen et al. 1990a, Jensen et al. 1990b). With the HUGIN approach efficient methods have been implemented for calculation of revised probability distributions for variables in a CPN without directed cycles (Andersen et al. 1989).

As always when modelling real world domains, the results infered from the model rely on the adequacy of the model and the reliability of data used. Therefore, no expert will blindly accept what the system comes up with. At least there will be kept a critical eye on the data, and mainly one will look for conflicts in the data or conflicts with the model.

In this paper we present a way of building such a critical eye into a system with a CPN model. Our suggestion requires an easy way of calculating probabilities for specific configurations. We start with a brief introduction to the HUGIN approach. In section 3 we discuss CPN's and data conflict. In section 4 a measure of conflict is defined, and it is shown that this measure is easy to calculate in HUGIN and that it supports a decomposition of global conflict into local conflicts. Section 5 reports on experience with a large CPN, and in section 6 we discuss how to distinguish between conflicts in data and data originating from a rare case.

## 2   Causal probabilistic Networks and the HUGIN approach

A causal probabilistic network (CPN) is constructed over a *universe*, consisting of a set of nodes each node having a finite set of *states*. The nodes are called *variables*. The universe is organized as a *directed acyclic graph*. The set of *parents* of $A$ is denoted by $pa(A)$. To each variable is attached a conditional probability table for $P(A|pa(A))$.

Let $V$ be a set of variables. The *space* of $V$ is the Cartesian product of the state sets of the elements in $V$ and is denoted by $Sp(V)$. The probabilitie tables are considered as functions and they are de-noted by greek letters $\phi$ and $\psi$. If $A$ is a variable, then $\phi_A = P(A|pa(A))$ maps $Sp(pa(A) \cup \{A\})$ into the unit interval $[0, 1]$. It is convenient to consider functions which are not normalized and take arbitrary non-negative values. So in the sequel, $\phi$ and $\psi$ denote such functions.

Evidence can by entered to a CPN in the form of *findings*. Usually a finding is a statement, that a certain variable is in a particular state.

After evidence has been entered to the CPN one should update the probabilities for the variables in the CPN. It would be preferable to have a local method sending messages to neighbours in the network. However, such methods do not exist when there are multiple paths in the network.

The HUGIN approach which is an extension of the work of Lauritzen and Spiegelhalter (1988) (Jensen et al 1990a; Jensen et al 1990b) represents one way of achieving a local propagation method also for CPN's with multiple paths. This is done by constructing a so-called junction tree which represents the same joint probability distribution as the CPN.

The nodes in a junction tree are *sets* of variables rather than single variables. Each node $V$ has a belief table $\phi_v : Sp(V) \rightarrow R_0$ attached to it. The pair $(V, \phi_v)$ is called a *belief universe*.

The crucial property of junction trees is that for any pair $(U, V)$ of nodes, all nodes on the path between $U$ and $V$ contain $U \cap V$.

A belief table is a (non-normalized) assessment of joint probabilities for a node. If $S \subset V$, then an (non-normalized) assessment of joint probabilities for $Sp(S)$ can be obtained from $\phi_V$ by *marginalization*: $\phi_S = \sum_{V \setminus S} \phi_V$.

Evidence can be transmitted between belief universes through the *absorption* operation: $(U, \phi_U)$ absorbs from $(V, \phi_V), \ldots, (W, \phi_W)$ by modifying $\phi_U$ with the functions $\sum_{V \setminus U} \phi_V, \ldots, \sum_{W \setminus U} \phi_W$. Actually, the new belief function $\phi'_U$ is defined as

$$\phi'_U = \phi'_U * \frac{\sum_{V \setminus U} \phi_V}{\sum_{U \setminus V} \phi_U} * \cdots * \frac{\sum_{W \setminus U} \phi_W}{\sum_{U \setminus W} \phi_U}$$

where the product $\phi * \psi$ is defined as

$$(\phi * \psi)(x) = \phi(x)\psi(x)$$

with $\phi$ and $\psi$ extended to the relevant space (if necessary).

Based on the local operation of absorption the two propagation operations *CollectEvidence* and *DistributeEvidence* are constructed. When CollectEvidence in $V$ is called (from a neighbour $W$) then $V$ calls CollectEvidence in all its neighbours (except $W$), and when they have finished their CollectEvidence, $V$ absorbs from them (see figure 1).

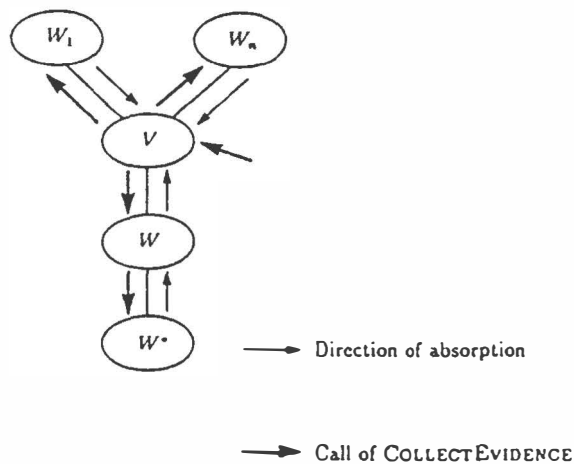

Figure 1: The calls and evidence passing in CollectEvidence

When DistributeEvidence is called in $V$ from a neighbour $W$ then $V$ absorbs from $W$ and calls DistributeEvidence in all its other neighbours.

Having constructed a junction tree, we need not be as restrictive with findings as in the case of CPN's:

Let $V$ be a belief universe in the junction tree. A *finding* on $V$ is a function

$$F_V : Sp(V) \rightarrow \{0, 1\}$$

So, a finding is a statement that some configurations of $Sp(V)$ are impossible. Note that the product of two findings $f : Sp(V) \rightarrow \{0, 1\}$ and $g : Sp(W) \rightarrow \{0, 1\}$ is a finding $f * g : Sp(V \cup W) \rightarrow \{0, 1\}$, and $f * g$ corresponds to the conjunction $f \wedge g$.

Using the HUGIN approach, it is possible to enter findings to the CPN (or the junction tree)[1], update the probabilities for all variables, and to achieve joint probability tables for all sets of variables which are subsets of nodes in the junction tree. The method has proved itself very efficient even for fairly large CPN's like MUNIN (see Olesen et al. 1989, Andersen et al. 1989).

[1] Actually, the more general notion of *likelihood* can be entered: *Evidence* is a function $Ev : Sp(V) \rightarrow R_0$. We will not pursue this in the present paper.

The main theorem behind the method is the following.

## Theorem 1

Let $T$ be any junction tree over the universe $U$, and let $\phi_U$ be the joint probability table for $U$.

(a) If CollectEvidence is evoked in any node $V$ and $\phi_V$ is the resulting belief table, then $\phi_V$ is proportional to $\sum_{U \setminus V} \phi_U$.

(b) If further, DistributeEvidence is evoked in $V$, then for any node $W$ the resulting belief table $\phi_W$ is proportional to $\sum_{U \setminus W} \phi_U$.

□

Before we proceed with data conflict, we will state an observation proved in Jensen et al. (1990b), but first noted by Lauritzen and Spiegelhalter (1988) in their reply to the discussion.

## Theorem 2

Let $T$ be a junction tree with all belief tables normalized, and let $x, \ldots, y$ be findings with prior joint probability $P(x * \ldots * y)$. Enter $x, \ldots, y$ to $T$ and activate CollectEvidence in any belief universe for $V$. Let $\phi_V^*$ be the resulting belief universe for $V$.

Then $\sum_V \phi_V^* = P(x * \ldots * y)$. □

## 3  CPN's and data conflict

A CPN represents a closed world with a finite set of variables and causal relations between them. These causal relations are not universal, but reflect relations under certain constraints. Take for example a diagnostic system which on the basis of blood analysis monitors pregnancy. Only diseases relevant for pregnant women are represented in the model. If the blood originates from a man, the constraints are not satisfied, and the *case is not covered by the model*. A similar situation appears if the test results are flawed (e. g. red herrings).

Following the tradition in probabilistic reasoning to take examples from California, where burglary and earthquake are everyday experiences, we have constructed the following example:

> When Mr. Holmes is at his office he frequently gets phone calls from his neighbour Dr. Watson telling him that his burglar alarm has gone off, and Mr. Holmes rushing home hears on the radio that there has been an earthquake nearby. Knowing that earthquakes have a tendency to cause false alarm, he then has returned to his office leaving his neighbours with the pleasure of the noise from the alarm. Mr. Holmes has now installed a seismometer in his house with a direct line to the office. The seismometer has three states:
>
> 0 for no vibrations
>
> 1 for small vibrations (caused by earthquakes or passing cars.)
>
> 2 for larger vibrations (caused by major earthquakes or persons walking around in the house.)
>
> The CPN for this alarm system is shown in figure 2:
>
> One afternoon Dr. Watson calls again and tells that the alarm has gone off. Mr. Holmes checks the seismometer, it is in state 0!

From our knowledge of the CPN, we would say that the two findings are in conflict. Performing an evidence propagation does not disclose that. The posterior probabilities are given in figure 3. Only in the rare situations of inconsistent data, an evidence propagation will show that something is wrong. The problem for Mr. Holmes is whether he should believe that the data originate from a rare case covered by the model, or he should reject that.

From a CPN model's point of view there is no difference between a case not covered by the model and flawed data. So what we can hope for to provide Mr. Holmes with is a measure indicating possible conflicts in the data given the CPN.

In MUNIN (Olesen et al. 1989) an attempt to incorporate conflict analysis in the CPN is made.

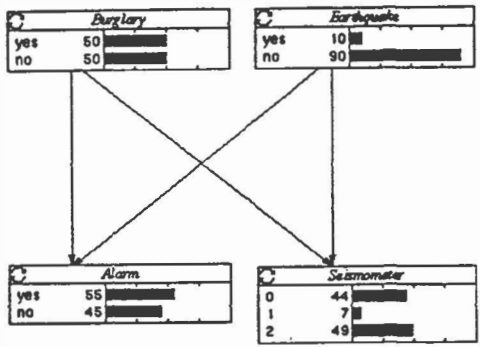

Burglary: $\phi_B : (50, 50)$; Earthquake: $\phi_E(90, 10)$

| $\phi_S$ | | $E$ | |
|---|---|---|---|
| | | N | Y |
| $B$ | N | (97, 2, 1) | (1, 97, 2) |
| | Y | (1, 2, 97) | (0, 3, 97) |

Seismometer

| $\phi_A$ | | $E$ | |
|---|---|---|---|
| | | N | Y |
| $B$ | N | (99, 1) | (1, 99) |
| | Y | (1, 99) | (0, 100) |

Alarm

Figure 2: Mr. Holmes' Alarm system with seismometer.

| $\phi_{E,B}$ | | $E$ | |
|---|---|---|---|
| | | N | Y |
| $B$ | N | .47 | .05 |
| | Y | .48 | 0 |

Figure 3: Joint probabilities for earthquake and burglary posterior to a : 'alarm = Y' and s : 'Seismometer = 0'.

This is done by introducing 'other'-states and 'other'-variables. In the example of Mr. Holmes' alarm system, an 'other'-variable covering lightening, flood, baseballs breaking windows etc. could be introduced to represent unknown causes for the alarm to go off, and the Burglar variable could have an 'other'-state covering Mr. Holmes' mistress having forgotten the code for switching off the burglary alarm.

Though this approach is claimed to be fairly successful, it raises several problems. First of all there is a modelling problem. The effect of an 'other'-statement is hard to model without knowing what 'other' actually stands for . What should the conditional probabilities be? In fact, these probabilities were in MUNIN constructed by feeding the network with conflicting data and thereby tuning the tables as to make 'other' light up appropriately.

A second problem is that conflict in data is a *global* property, and the introduction of 'other'-statements in the CPN gives only a possibility of evaluating evidence locally. In order to combine the local 'other' statements to a global one, the CPN has to be extended drastically.

This leads to the third major problem, which is more of a technical kind. The introduction of 'other'-statements to the CPN can cause a dramatic increase in the size of the junction tree. Besides, the technique with 'other'-states is hard to use if the variables are not discrete.

Another approach has been suggested by Habbena (1976). It consists of calculating a *surprise index* for the set of findings. Essentially, the surprise index of $f : V \to \{0, 1\}$ is the sum of the probabilities of all findings on $V$ with probabilities no higher than $f$'s.

Habbena suggests that a threshold between 1% and 10% should be realistic. In the seismometer case, the surprise index for $(a, s)$ is 3%. However, the calculation of a surprise index is exponential in the number variables in $V$ and must be considered as intractable in general.

## 4 The conflict measure conf

Our approach to the problem is that correct findings originating from a coherent case covered by the model should conform to certain expected patterns. If $x, \cdots, y$ are the findings, we therefore

should expect:

$$P(x * \cdots * y) > P(x) \times \cdots \times P(y)$$

Hence we define the conflict measure conf as:

$$\text{conf}(x, \cdots, y) = \log \frac{P(x) \times \cdots \times P(y)}{P(x * \cdots * y)}$$

(where log is with base 2).

This means that a positive $\text{conf}(x, \cdots, y)$ is an indicator of a possible conflict.

For the data in section 3 we have $\text{conf}(a, s) = 4.5$.

Using theorem 2, $\text{conf}(x, \cdots, y)$ is very easy to calculate in HUGIN. The prior probabilities $P(x), \cdots, P(y)$ are available before the findings are entered, and $P(x, \cdots, y)$ is the ratio between the prior and the posterior normalizing constant for any belief universe.

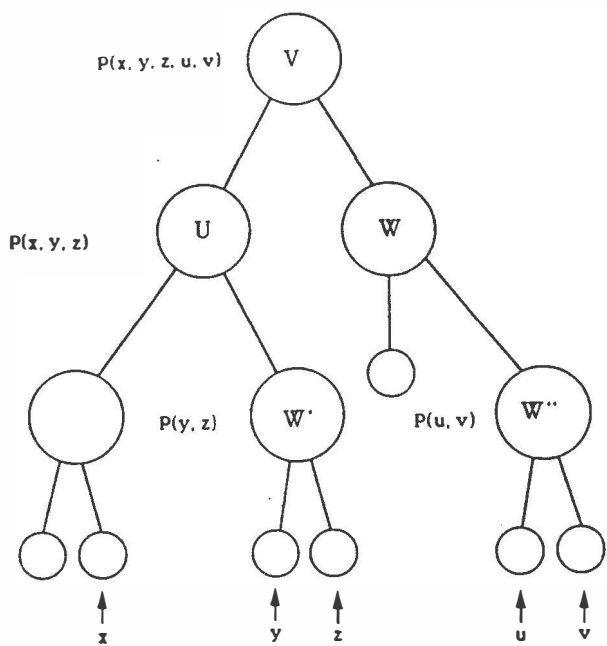

Figure 4: A junction tree with findings $x, y, z, u, v$ entered. Theorem 2 provides the joint probabilities indicated at nodes V, U, W' and W''.

The conflict analysis can be further refined. In figure 4 is shown a junction tree with findings $x, y, z, u, v$ entered. If CollectEvidence is evoked in the node $V$, then the evidence flowing to V consists of two sets of findings, namely $\{x, y, z\}$ and $\{u, v\}$. Since the product of findings is also a finding, we can say that the two findings $x * y * z$ and $u * v$ meet in V.

The conflict in the data meeting in V is therefore composed of the conflict between $x * y * z$ and $u * v$, the conflict inside $\{x, y, z\}$ and inside $\{u, v\}$. It is easy to show that:

$$\text{conf}(x, y, z, u, v) = \text{conf}(x * y * z, u * z)$$
$$+\text{conf}(x, y, z) + \text{conf}(u, v)$$

Furthermore, as indicated at figure 4, $P(x * y * z)$ and $P(u * v)$ can be calculated as ratios between prior and posterior normalizing constants, and therefore $\text{conf}(x, y, z)$ and $\text{conf}(u, v)$ as well as $\text{conf}(x * y * z, u * z)$ are easy to calculate.

In general: If evidence is propagated to any belief universe U from neighbours $V, \cdots, W$ originating from findings $(v, \cdots v') \ldots (w, \cdots, w')$ respectively, then

$$\text{conf}(v, \cdots, v', \cdots, w, \cdots w') =$$
$$\text{conf}(v * \cdots * v', \cdots, w * \cdots * w')$$
$$+\text{conf}(v, \cdots, v') + \cdots + \text{conf}(w, \cdots, w')$$

All terms are in HUGIN easy to calculate by use of Theorem 2.

We call $\text{conf}(v, \cdots, v', \cdots, w, \cdots, w')$ the *global* conflict and $\text{conf}(v * \cdots * v', \cdots, w * \cdots * w')$ the *local* conflict.

The calculation of conf has been implemented in HUGIN to follow the calls of CollectEvidence. The overhead to the propagation methods in terms of time and space is neglectable.

## 5  Example: APB-MUNIN

The conflict measure has been tested on small fictious examples and on a large subnetwork of MUNIN, namely the network for the muscle Abductor Pollicis Brevis (APB). The network is shown in figure 5.

The rightmost variables in figure 5 are *finding variables*. This means that evidence is entered at the right hand side of the CPN and propagates to the left. However, as described in section 2, the propagation takes place in a junction tree of belief universes. In the test, CollectEvidence was called in universe number **59**, and the call propagates recursively down the junction tree. In figure 6 is shown the junction tree. (Only belief universes where evidence meet are shown).

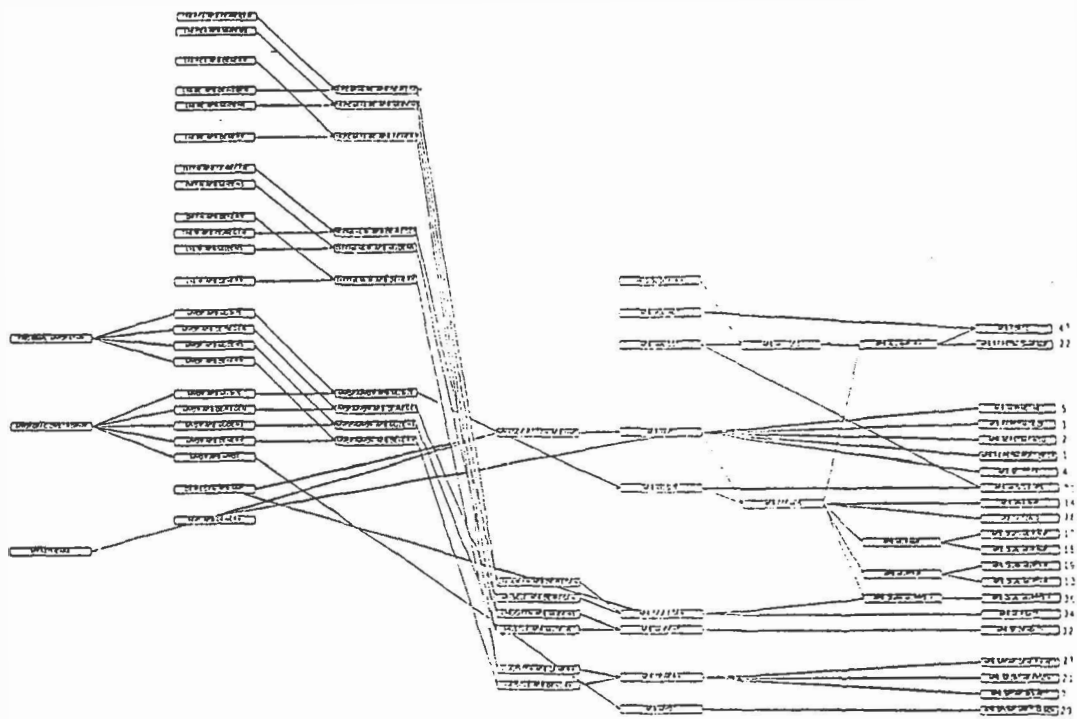

Figure 5: The DAG in MUNIN for *Medianus Ab-ductor Pollicis Brevis.* The attached numbers indicate the belief universe to which the finding is entered (see figure 6.)

First we asked the model builder, Steen Andreassen, to provide us with a complete set of normal findings. They were entered, and global and universal conflict values were calculated. The results are shown in figure 7. Surprisingly we got a global conflict of 23.3 for the entire set of findings and apparently the conflict can be traced to belief universe no. **45**. Further, the evidence from **15** and **17** looks conflicting. Returning to Steen Andreassen with our surprise, he recognized that he had given us a wrong value for the finding qual.mup.amp. which was entered to belief universe **15**. It should have been 540 $\mu V$ rather than 200 $\mu V$.

We entered the corrected finding and got a global conflict value -1.5 for the entire set of findings with local and subglobal values ranging between 0 and -1.4.

Then typical findings for a patient suffering from moderate proximal myopathy were entered. As can be seen in figure 8, this resulted in large

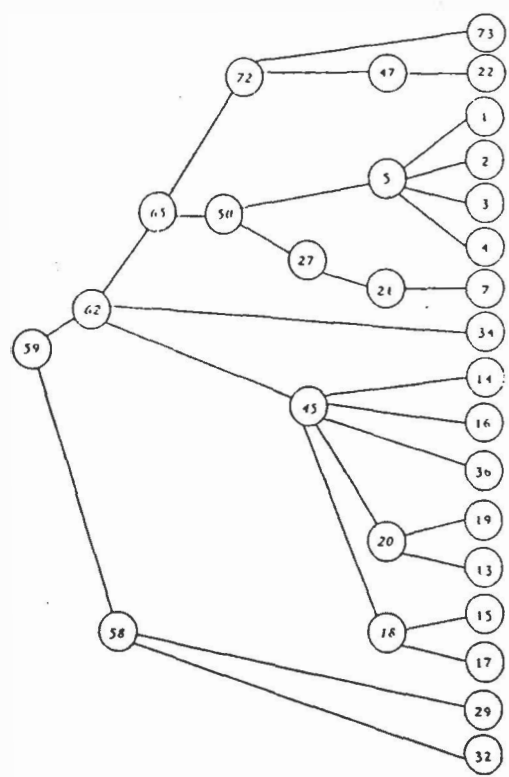

Figure 6: The part of the MUNIN junction tree for APB where evidence meet. The numbers are labels of belief universes. Bold numbers indicate entrance of findings.

negative conflict values confirming the coherence of the findings.

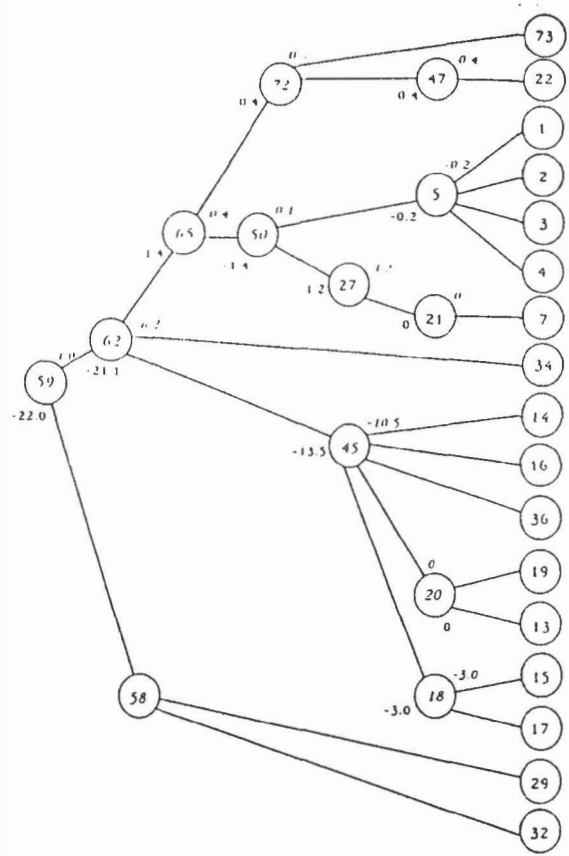

Figure 8: Typical findings for a patient suffering from moderate proximal myopathy entered.

Finally, we simulated hypothesizing. We entered a set of findings originating from a healthy patient, and we also entered the disease state 'moderate proximal myopathy'. The result is shown in figure 9. The disease finding is entered to belief universe **58**, and it can be seen that the disease does not contradict a couple of normal findings, but indeed the whole set.

## 6 Conflict or rare case?

It can happen that typical data from a very rare case might cause a high value of conf. In the case of Mr. Holmes' alarm system a flood (with probability $10^{-3}$ could be entered to the CPN explaining the data (see figure 10).

For this system we get $\text{conf}(a, s) = 4.5$. It is still indicating a possible conflict. The reason is that though $P(a, s)$ is possible, it is under the rare condition of flood. Mr. Holmes looks out of the window. It rains cats and dogs, and he has resolved the problem; the model gives a new $P(\text{Flood}) = 0.84$.

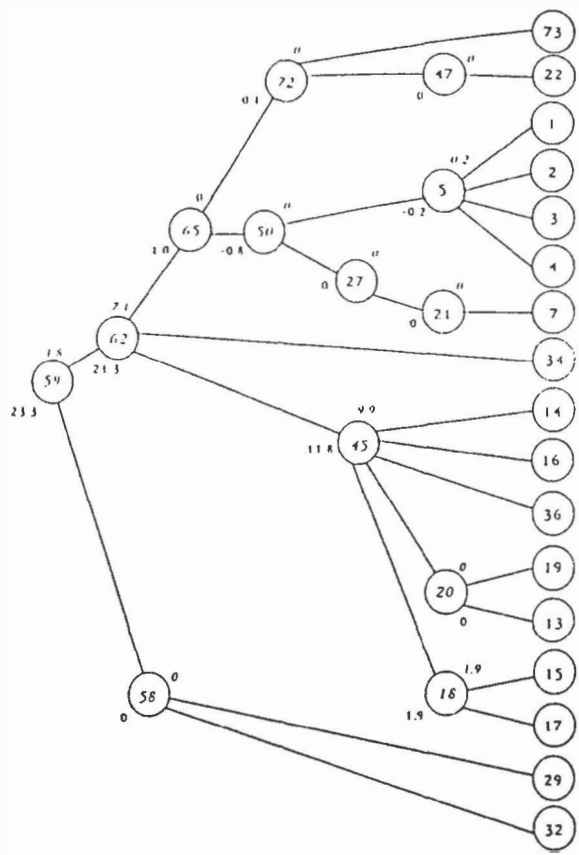

Figure 7: The conflict measures from the first test example. The italiced values are local conflict values and the bold figures are the global ones.

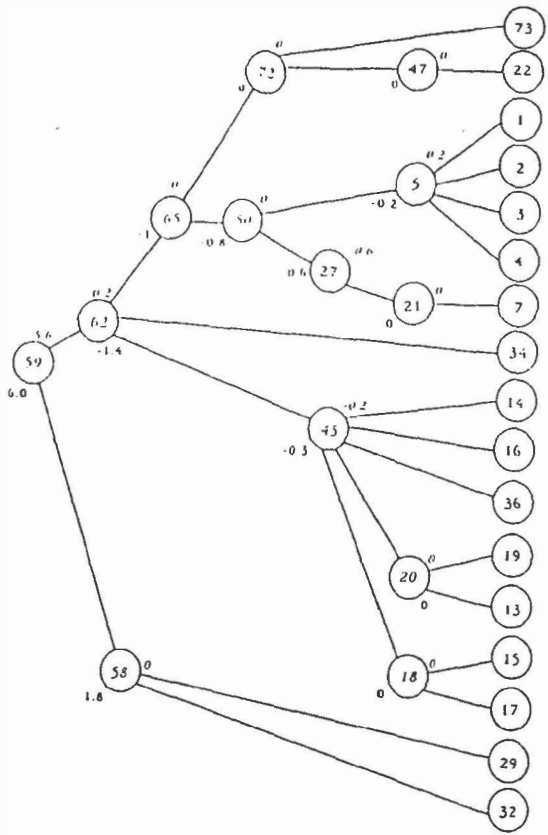

Figure 9: Findings for a healthy patient, and the hypothesis 'moderate proximal myopathy' entered.

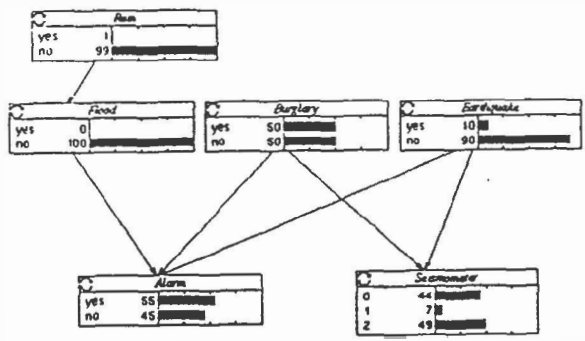

Figure 10: Mr. Holmes' revised CPN.

The problem above calls for more than a possibility for refined conflict analysis. We need a method to point out whether a conflict can be explained away through a rare cause.

Let $(x, \ldots, y)$ be findings with a positive conflict measure, and let $H$ be a hypothesis which could explain the findings: $\text{conf}(x, \ldots, y, H) < 0$
We have

$$\text{conf}(x, \ldots, y, H) = \log \frac{P(x) \times \ldots \times P(y) \times P(H)}{P(x * \ldots * y * H)}$$

$$= \text{conf}(x, \ldots, y) + \log \frac{P(H)}{P(H|x, \ldots, y)}$$

This means that if

$$\log \frac{P(H|x, \ldots, y)}{P(H)} > \text{conf}(x, \ldots, y)$$

then $H$ can explain away the conflict.

The left-hand ratio can be monitored automatically for all variables (in the flood example the value is 5.6). This means that there is no need for manually to formulate explaining hypothesis in terms of states of variables. More complex hypothesis can also be monitored if they can be expressed as findings.

## 7 Conclusion

The measure of conflict

$$\text{conf}(x, \cdots, y) = \log \frac{P(x) \times \cdots \times P(y)}{P(x * \cdots * y)}$$

has many promising properties. It is easy to calculate in HUGIN, it is independent of the order in which findings are entered, it can be used for both global and local analysis of conflicts in data, and it has a natural interpretation which supports the usual mental way of inspecting data for flaws or for originating from sources outside the scope of the current investigation.

However, still some practical and theoretical work is needed in order to understand the significance of specific positive conflict values. Also, the detailed conflict analysis a it is nowconnected to the structure of the junction tree rather than to the CPN itself. This should be relaxed.

## 8 Acknowledgements

We thank Steffen Lauritzen for many valuable discussions on the subjects of this paper, and Steen Andreassen for helping with the MUNIN experiment.